\documentclass{article}[10pt]
\usepackage{tikz}
\usepackage{xcolor}
\usepackage{amsmath}
\usepackage{mathdots}
\usetikzlibrary{math, positioning, decorations.pathreplacing, backgrounds, decorations.shapes}
\usepackage[dvipsnames]{xcolor}
\usepackage[numbers]{natbib}
\usepackage{graphicx}
\usepackage{hyperref}
\usepackage{enumitem}
\usepackage{ stmaryrd }
\usepackage{subfigure}
\usepackage{float}

\usepackage{amsmath}
\usepackage{amsthm}
\usepackage{amssymb}

\usepackage{booktabs}
\usepackage{xspace}
\usepackage[linesnumbered, ruled, vlined, noend]{algorithm2e}
\usepackage[capitalize]{cleveref}

\usepackage{microtype}

\theoremstyle{plain}

\newtheorem{remark}{Remark}

\newcommand{\param}{\textsc{ParamExplorer}\xspace}
\newcommand{\processing}{\textsc{Processing}\xspace}
\newcommand{\pjs}{\textsc{p5$.$js}\xspace}
\newcommand{\cmaes}{CMA-ES\xspace}

\title{\param: A framework for exploring parameters in generative art}

\author{
    Julien Gachadoat\thanks{Generative artist, Bordeaux, France. \url{https://www.v3ga.net/}} \and
    Guillaume Lagarde\thanks{LaBRI, University of Bordeaux, France. \url{https://guillaume-lagarde.github.io/}}
}

\begin{document}
\maketitle

\begin{abstract}
Generative art systems often involve high-dimensional and complex parameter spaces in which aesthetically compelling outputs occupy only small, fragmented regions. Because of this combinatorial explosion, artists typically rely on extensive manual trial-and-error, leaving many potentially interesting configurations undiscovered. In this work we make two contributions. First, we introduce \param, an interactive and modular framework inspired by reinforcement learning that helps the exploration of parameter spaces in generative art algorithms, guided by human-in-the-loop or even automated feedback. The framework also integrates seamlessly with existing \pjs projects. Second, within this framework we implement and evaluate several exploration strategies, referred to as agents.
\end{abstract}

\section{Introduction}\label{sec:intro}

\emph{Generative art} refers to artistic creation driven by automated procedures specified through a set of formal rules. Initially known as \emph{computer art}, the practice emerged in the post--World War~II period with the advent of early computational technologies. By the 1960s and 1970s, a first generation of engineer-artists had begun to appropriate these machines for creative purposes, treating algorithms as a novel artistic medium and opening avenues for the exploration of new visual forms. Computers, tireless and precise, quickly became valuable partners for artists working with code. For detailed historical overviews, see~\cite{dreher2014history, Migayrou2018CoderLeMonde}.

In the early 2000s, access to generative art expanded significantly. The rise of the internet, together with the introduction of tools such as John Maeda's \emph{Design by Numbers}~\cite{maeda2001design} and the \emph{Processing} environment developed by Casey Reas and Ben Fry~\cite{reas2007processing, processing2001}, helped introduce this culture of creative coding to a much wider audience. These simple and approachable tools allowed people with little or no programming experience to write their own interactive digital works.

The starting point of this paper lies in a recurring challenge in algorithmic and generative art: the exploration of the parameter spaces underlying generative processes. In such systems, \emph{parameters} are the adjustable variables that modify the behavior of the algorithm and change the resulting outputs. These spaces are often high-dimensional, nonlinear, and fragmented into disjoint regions that produce qualitatively different outputs. In this landscape, only a small subset of these regions tends to yield aesthetically interesting results, making the search for compelling configurations akin to finding a needle in a haystack.

A typical generative art algorithm may expose dozens of parameters, each influencing a distinct aspect of the generated images. Because the volume of the search space grows exponentially with dimensionality, systematic exploration quickly becomes infeasible, a phenomenon known as the \emph{curse of dimensionality}. In practice, artists therefore rely heavily on intuition and incremental adjustments to identify promising parameters. Vera Molnár already described this process, with its advantages and limitations, as early as 1975:

\begin{quote}
“When I have an idea for a new picture, I make
the first version of it rather quickly. Usually I am
more or less dissatisfied with it and I modify it.
I alter in a stepwise manner the dimensions, proportions
and arrangement of the shapes. When simple geometrical
shapes are used, such modifications are relatively easy to make.
By comparing the successive pictures resulting from a series of
modifications, I can decide whether the trend is toward the result
that I desire. What is so thrilling to experience is not only the
stepwise approach toward the envisioned goal but also sometimes
the transformation of an indifferent version into one that I find
aesthetically appealing.

This stepwise procedure has, however, two important disadvantages
if carried out by hand. Above all, it is tedious and slow. In order to
make the necessary comparisons in a developing series of pictures,
I must make many similar ones of the same size and with the same
technique and precision. Another disadvantage is that, since time
is limited, I can consider only a few of many possible modifications.
Furthermore, these choices are influenced by disparate factors such
as personal whim, cultural and educational background and ease of
execution.”

\hfill\textemdash\ Vera Molnár, \emph{1975}~\cite{Molnar1975}
\end{quote}

While this exploratory process can be rewarding, it is both time-consuming and subject to cognitive biases. In particular, artists often gravitate toward familiar stylistic patterns, which can lead to under-exploration of unfamiliar yet potentially interesting areas of the parameter space. Indeed, many compelling outputs are discovered only by chance, while randomly sampling parameters, suggesting that an automatic guided exploration could reveal more efficiently creative opportunities. This is what we aim to address in this work.

\subsection*{The feedback loop in artistic exploration}

The typical creative workflow of generative art can be viewed as a feedback loop between the artist and its generative system. The artist observes a produced image, evaluates its aesthetic, and decides whether to refine the current parameter configuration, to move to a different region of the parameter space or even make small modifications to the generative algorithm itself. This iterative process closely resembles reinforcement learning (RL)\footnote{More precisely, it can be viewed as a multi-armed bandit problem, one of the simplest forms of RL, also known as a markov decision process with one state.}: the artist acts as an agent providing an action (a value for every parameters of the system, or modifying the code), while the generative algorithm functions as the environment that, in response to this action, produces an observation in the form of a generated image. The artist then evaluates this output and provides feedback, which can be interpreted as a reward signal guiding future actions. Over time, through repeated interaction, the artist learns to navigate the parameter space more effectively, gravitating toward regions that yield aesthetically satisfying results.

This perspective motivates the design of computational tools that formalize and support this exploratory process. Instead of requiring the artist to manually specify new parameter values at every iteration, RL-inspired techniques can help automatically propose promising next actions based on previously received feedback. However, unlike traditional RL settings, the objective here is not to converge to a single optimal solution. Rather, the goal is to uncover a diverse set of high-quality outputs, reflecting the inherently exploratory and open-ended nature of creative practice. Achieving this requires methods that balance exploitation of known promising regions with exploration of underexplored areas of the parameter space.

\subsection*{\processing and \pjs}

We focus on the \processing ecosystem and \pjs because of their widespread adoption among generative artists and creative coders. \processing~\cite{reas2007processing, processing2001}, created by Casey Reas and Ben Fry in 2001, is both a software environment and a programming language designed to teach coding within the context of the visual arts, and it has been widely embraced by the creative coding community. Its JavaScript counterpart, \pjs~\cite{mcCarthyp5js}, extends these capabilities to web-based environments, making it easier to share online.

Our goal is to develop a framework in which existing \pjs projects can be imported directly, without requiring modifications to the original code, in order to enable artists to quickly experiment with different exploration strategies on their own generative systems.

\subsection*{Our contributions}
This work makes two main contributions:
\begin{itemize}
    \item We introduce \param, an interactive and modular framework inspired by reinforcement learning that facilitates the exploration of high-dimensional parameter spaces in generative art algorithms. The framework accommodates both human-in-the-loop and automated aesthetic feedback, and is designed to integrate seamlessly with existing \pjs projects.
    \item Within this framework we design and evaluate several exploration strategies (referred to as \emph{agents}).
\end{itemize}

\begin{remark}\label{rem:manual-exploration}
    Although the primary goal of this work is to facilitate and test artistic exploration through agents, the framework can also be used for fully manual exploration of parameter spaces. Its built-in database stores generated images along with their associated parameters and scores, enabling users to track and visualize their exploration history through the gallery, and to revisit or refine previous configurations.
\end{remark}

\subsection*{Related work}

Within the general problem of exploring complex, high-dimensional landscapes, a closely related line of work is black-box hyperparameter optimization (HPO), where the goal is to automatically find optimal hyperparameter settings for machine learning algorithms, such as learning rate, regularization strength and architecture design, to name a few. This problem shares similarities with our setting, as it requires searching for good parameter configurations within complex, high-dimensional spaces. Many approaches have been proposed in HPO, including Bayesian optimization, bandit algorithms, and evolutionary strategies such as \cmaes; see, for example, the survey by~\cite{feurer2019hyperparameter} and the Optuna toolkit~\cite{akiba2019optuna}. While our framework inherits HPO's challenge of navigating multimodal landscapes, it differs in at least two aspects. First, feedback in our context is subjective and noisy\footnote{Humans naturally struggle to consistently rank visual artworks, whose aesthetics do not really admit a clear total ordering.}, as it stems from human aesthetic judgment rather than an objective performance metric. Second, whereas HPO aims to identify a single optimal configuration, our goal is to uncover a diverse set of aesthetically interesting outcomes. This emphasis on diversity makes standard HPO strategies insufficient on their own and motivates the integration of niching and exploration-focused techniques within our framework.

More broadly, these challenges fall under the umbrella of multi-modal optimization (MMO), in which the objective is not to identify a single optimum but to discover multiple high-quality solutions distributed across a complex search landscape; see for example~\cite{li2016seeking} for a comprehensive survey. This formulation naturally aligns with creative exploration, where the goal is to uncover diverse and aesthetically interesting regions rather than converge to a single best outcome.

An interesting and closely related work is that of McCormack and Gambardella~\cite{mccormack2022quality}, who apply quality-diversity search to explore a specific line-based generative system for physical plots. Their approach differs from ours in two main respects. First, their method operates in the image space using learned feature embeddings, whereas our agents explore the underlying parameter space. Second, their process unfolds in two separate stages: a fitness measure is first learned from user feedback on a hundred images using a neural network, after which a quality-diversity algorithm autonomously explores the space without further human input. By contrast, our framework supports a continuous and never-ending feedback loop between the agent and the user. Finally, McCormack and Gambardella do not aim to propose a generic framework. Indeed, the agent they developed could be integrated into \param as an additional exploration strategy, making their approach complementary rather than competing.

In a similar line of work, where an agent assists the artist, we can also cite Ibarrola et al.~\cite{ibarrola2022towards}, who propose an agent to co-create drawings using Contrastive Language–Image Pretraining (CLIP), but their system focuses on interactive, gesture-based image manipulation rather than on a feedback loop for exploring the parameter space of a generative algorithm.

\section{Framework overview}\label{sec:framework}

For more information about the framework, we refer to the GitHub repository of the project~\url{https://github.com/guillaume-lagarde/param-explorer}.

The proposed framework has two primary objectives.  
First, it enables the seamless integration of existing \texttt{p5.js} projects (and, in the future, other creative coding frameworks) into an exploration environment without requiring any refactoring from the artist or developer.  
Second, it provides a flexible architecture for implementing new exploration strategies (the agents), in either JavaScript or Python.

In addition, the framework can be used as a standalone tool for manual exploration of parameter spaces, as explained in Remark~\ref{rem:manual-exploration}.

We now describe the main components of the framework and how they interact.

\subsection*{Choice of parameters to explore}

Once the algorithm is imported, the user must first select which parameters of the generative system they wish to explore with \param, along with their respective ranges. This means selecting a subset of the variables defined in the \pjs sketch that will be exposed to the exploration framework. In practice, parameters worth exposing are those that significantly influence the visual output. The bounds of these parameters are determined using the artist's intuition, aiming for ranges that are both aesthetically meaningful and computationally tractable\footnote{For example, a recursion depth parameter might be clipped to some reasonable maximum value to avoid an unreasonable running time.}.

\subsection*{Agent-User interaction loop}

The agent understands the language of parameter space, while the user understands the language of image space. Communication between the two is made possible because the image space often varies smoothly enough with respect to the underlying parameter space.

\begin{figure}[H]
    \centering
    \includegraphics[width=\columnwidth]{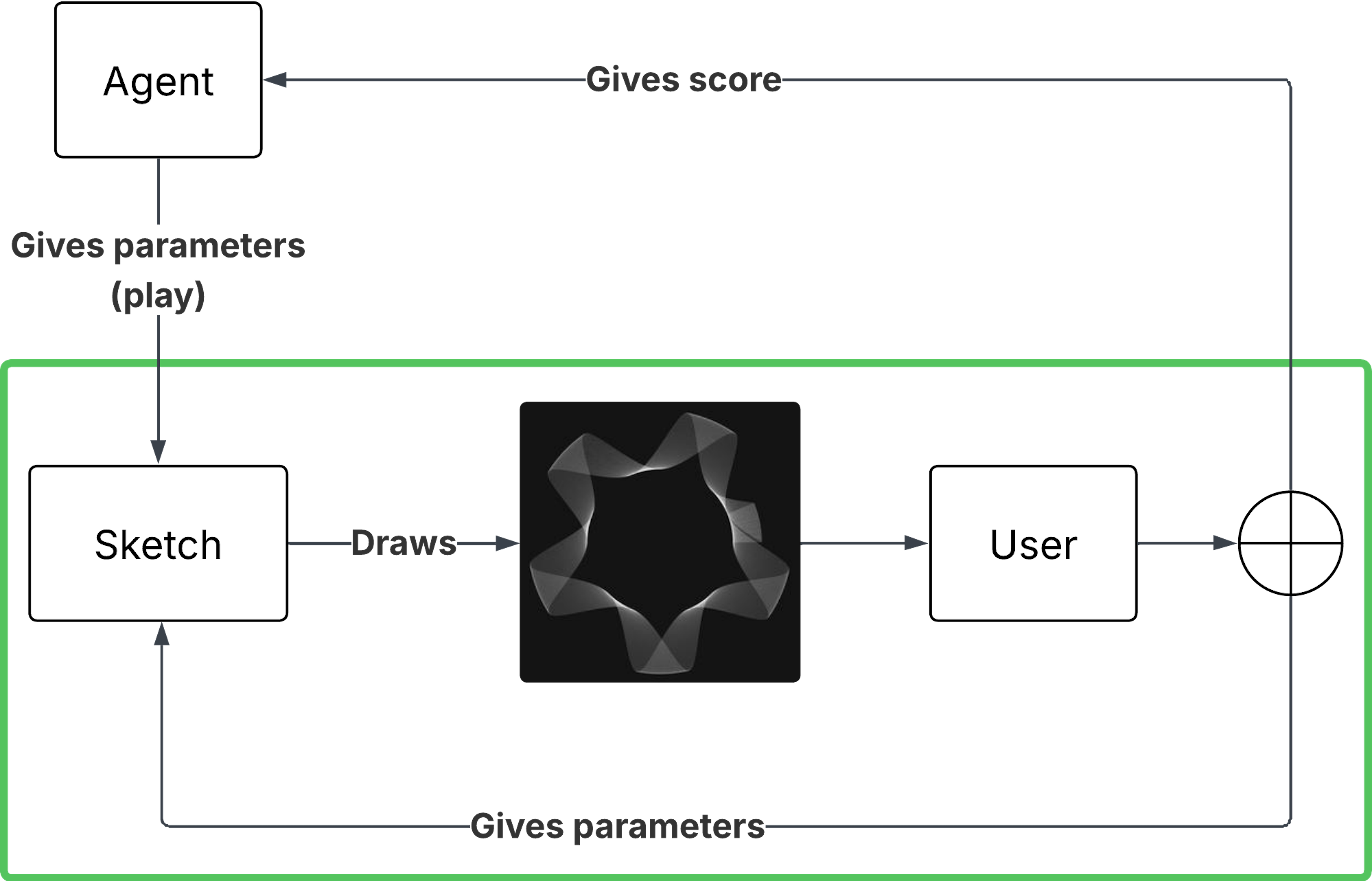}
    \caption{Interaction loop between the user and the agent. 
    The green box represents the traditional manual feedback loop between the 
    user and the generative system. The agent extends this loop and gives the 
    user two additional options: (1) provide feedback to the agent, which uses 
    the score to update its internal state and propose new parameters; and (2) refine the current drawing 
    proposed by the agent by manually adjusting the parameters. 
    As before, the user can also generate a new image manually.
    Additionally, the user may give instructions to the agent, such as moving forward or backward in its internal timeline to help guide its exploration-exploitation balance.}
    \label{fig:interaction_loop}
\end{figure}

Within this loop, the agent proposes a set of parameters, which is transformed into a drawing through the user's \pjs sketch. The user then inspects the generated image and provides feedback by assigning a score. This score is then sent back to the agent to update its internal state. See Figure~\ref{fig:interaction_loop} for an illustration of this interaction loop.

\subsection*{Workflow in practice}

The interaction loop is centred on user visual evaluations of batches of generated images. The typical workflow proceeds as follows:
\begin{enumerate}
    \item The user runs a \emph{batch generation}. The selected agent proposes several parameter configurations, which are rendered on the canvas.
    \item The user inspects the results and \emph{assigns scores} to some of the images.
    \item These scores are transmitted to the agent, which updates its internal model based on these new feedbacks.
    \item At any moment, the user may invoke the \textit{time\_warp} function to move forward or backward in the agent's internal timeline. Stepping forward makes the agent rely more strongly on previously successful regions of the parameter space; stepping backward restores a more exploratory behaviour. In practice, if the current batches lack diversity, the user might want to step backward to recover broader exploration. Conversely, when the user wants to refine a promising family of outputs, stepping forward narrows the search around the agent's most recent estimates.
\end{enumerate}

This cycle can be repeated as many times as desired.

\subsection*{Framework structure}

The system relies on three main components organized in a client-server architecture :
\begin{itemize}
    \item a front-end tool handling rendering, user interface controls, batch orchestration, and interactive tools.
    \item a back-end agent infrastructure.
    \item a database in tinydb format to store a history of generated images, their parameters, user scores, etc. 
\end{itemize}

Having chosen \pjs as the graphics library, the front-end tool was naturally developed as a web application using HTML, CSS, and JavaScript. Using the browser as a platform offers us maximum compatibility across different operating systems and simplified distribution by uploading it to a server. 
The tool integrates a library of graphic components that Julien Gachadoat has been maintaining and using since 2021 in the development of his own tools. The application communicates with the server via HTTP requests (\texttt{fetch}) and messages in JSON format.
The tool has three main tabs: 
\begin{itemize}
        \item The \emph{“Manual exploration”} tab allows users to generate random drawings, edit drawings saved in the database, and rate these drawings. This tab displays a graphical interface automatically created by the tool based on the parameters specified in \param.
        \item The \emph{“Agent exploration”} tab allows users to generate drawings using a selected agent. The tab works in batches, which is useful for quickly creating a set of drawings. This tab becomes especially relevant when agents have already been trained on a series of drawings rated by the user, but it can also be used in cold start mode (see the paragraph on cold start for a word on this).
        \item The \emph{“Gallery”} tab shows all the generated images, they can be organized according to several predefined filters. The interface allows users to select drawings by group in order to rate them (and thus train agents) or delete them.
\end{itemize}

The server was written in Python and manages the tool's HTTP requests. It allows drawings (parameters, notes, and images) to be saved and the various agents to be trained. Python was the natural choice for us due to its abundance of libraries in the field of mathematics and statistics, which facilitated the development of agents. 

The server communicates directly with a tinydb-type database, which saves information about the generated images (parameters, path to the image on the disk) as well as the agents (internal state). This architecture is lightweight and flexible and allows us to easily sort images according to their score or timestamp, for example.

\subsection*{Cold start and manual exploration}\label{subsec:cold-start}

The framework also supports manual exploration of the parameter space, without invoking any agent. This is particularly useful at the beginning of the process when the agent has no data, a phenomenon known as \emph{cold start}: indeed the user can freely choose parameter configurations, score the corresponding outputs, and build an initial set of evaluations before switching to an automated strategy. Alternatively, the user may let an agent generate one or several random batches to obtain a first pool of scored examples.

Manual exploration remains useful later in the workflow as well. It allows the user to refine specific regions of the parameter space, including those revealed by the agent. These targeted adjustments can then be fed back to update the agent, helping it focus its next proposals closer to the current promising area.

\section{Agent design}\label{sec:agents}

Agents are responsible for proposing new set of promising parameters and updating their internal state based on user feedback.
Implementing a new agent essentially consists in defining three main functions:

\begin{itemize}
    \item \texttt{play() $\rightarrow$ dict}: generates and returns a new set of parameters to evaluate;
    \item \texttt{update(parameters: dict, score: float, metadata: dict) $\rightarrow$ None}: updates the agent's internal state based on a set of parameters, an associated score given by the user (or an automated process), and any additional metadata;
    \item optionally, \texttt{time\_warp(steps: int) $\rightarrow$ None}: shifts the agent's internal state forward or backward in time.
\end{itemize}

Three different agents are currently implemented in the framework, we describe them below.

\subsection*{Random agent}
This agent serves as a baseline. Its \texttt{play} function samples parameter configurations uniformly at random within the user-defined bounds, and its \texttt{update} function performs no operation: it does not learn from feedback. Despite its simplicity, this strategy remains useful, as it mirrors a common exploratory practice among generative artists.

\subsection*{Gaussian agent}
This agent is designed to mimic how a human artist explores a generative system. It combines two main ideas: local exploration around previously evaluated parameter configurations that received good scores, performed via Gaussian sampling; and a clustering-based mechanism to preserve diversity in the search process.

It maintains a small portfolio $\mathcal{P}$ of previously seen parameter configurations (typically those that the user found interesting) each associated with one of $k$ internal "modes" (the population indices). Each mode $i$ has its own standard deviation $\sigma_i$, which controls the exploration radius around that mode and decays over time as the agent receives feedback. The hyperparameter $k$ determines how many distinct regions of the parameter space the agent is encouraged to explore: a large $k$ promotes diversity, while a smaller $k$ leads to more concentrated search. It also has a decay factor $\gamma \in (0,1)$ to control how quickly the exploration radii shrink over time.

At each \texttt{play} call, the agent proceeds as follows:
\begin{enumerate}
\item If the size of the portfolio is less than $k$ (warm-up phase), it samples a parameter configuration uniformly at random and assigns it a new mode.
\item Otherwise, it selects one previously stored configuration from $\mathcal{P}$ uniformly at random, along with its associated mode index $i$.
\item It samples a new parameter configuration from a Gaussian distribution centered on the selected configuration, using that mode's current standard deviation $\sigma_i$.
\end{enumerate}

During a call to \texttt{update}, the agent stores the new set of parameters and its associated score and mode $i$ from which they originated. It then decays $\sigma_i$ multiplicatively by $\gamma$ to focus exploration within that mode. When the portfolio becomes too large, the agent clusters all stored configurations into $k$ groups and retains only the best-scoring individual from each cluster. During this reduction step, if a cluster contains only a few points, we increase its associated $\sigma_i$, as this suggests that the corresponding local region of the search space is not promising\footnote{The intuition is that if the region were promising, the user would have added more points from that cluster into the portfolio.} and thus more exploration is needed for that particular mode. Overall, this strategy preserves $k$ diverse modes while preventing any single mode from dominating the portfolio.

The \texttt{time\_warp(steps)} function applies the same multiplicative decay used during updates to each mode's standard deviation. Concretely, for every mode $i$, it performs
\[
    \sigma_i \leftarrow \sigma_i \cdot \gamma^{\text{steps}},
\]
thereby shrinking (or expanding, if \texttt{steps} is negative) all exploration radii as if \texttt{steps} regular updates had occurred.
\subsection*{Open-ended agent}

This agent is similar to the Gaussian agent but uses an open-ended population mechanism designed to continually expand the search space whenever the user provides positive feedback or manually contributes promising configurations. Unlike a fixed-population evolutionary strategy, this agent dynamically grows its set of search "modes", each represented by a population index, whenever it encounters a configuration deemed worth exploring further.

The agent maintains a collection of populations where population of index $0$ is reserved exclusively for uniform exploration. Every other population index corresponds to a \textit{discovered region} of the parameter space, each storing a small history of $(\text{parameters}, \text{score})$ pairs and equipped with an exploration radius $\sigma_i$. These radii decay over time as feedback is incorporated to shift toward fine-grained refinement. A global list of repulsive points prevents the agent from repeatedly sampling known unproductive areas.

At each \texttt{play} call, the agent proceeds as follows:
\begin{enumerate}
    \item It selects a population index $i$ uniformly at random.
    \item If $i = 0$, it samples a parameter configuration uniformly from the parameter space, avoiding previously stored repulsive points.
    \item If $i > 0$, it picks a configuration from that population's history and samples new parameters from a Gaussian distribution centered at that configuration with radius $\sigma_i$.
\end{enumerate}

The \texttt{update} function is similar to that of the Gaussian agent, except with three main differences:
\begin{itemize}
    \item when the population index is $0$, it creates a new population with the provided parameters and score
    \item when the score is zero, it adds the parameters to the repulsive set.
    \item when the history of a population exceeds its maximum length, it reduces it by subsampling representative elements.
\end{itemize}

The \texttt{time\_warp(steps)} function is identical to that of the Gaussian agent.

\subsection*{\cmaes agent}

This agent implements a multi-population variant of the Covariance Matrix Adaptation Evolution Strategy (\cmaes), coupled with a simple niching mechanism.\footnote{Many niching strategies have been proposed in the multi-modal optimization community; see, for example,~\cite{li2016seeking} for a survey of such methods. In this paper we opt for a niching strategy that is easy to implement and let future work explore more sophisticated alternatives.} 
The main conceptual difference compared to the previous agents is that \cmaes learns correlations between parameters and adapts its search distribution accordingly, whereas the earlier agents rely on isotropic Gaussian sampling. For example, \cmaes can detect when varying two parameters jointly leads to better outcomes and will assign them a strong correlation.
It maintains $n$ independent \cmaes sub-populations (or "niches"), each with its own mean parameter vector $m_j$, global step-size $\sigma_j$, and covariance matrix $C_j$. Each sub-population explores a different region of the parameter space, and if two of them collapse onto one another, one is restarted to preserve diversity.

At each \texttt{play} call, the agent proceeds as follows:
\begin{enumerate}
    \item It selects one of the $n$ sub-populations: during the first $n$ calls each is used once, and afterwards one is chosen uniformly at random.
    \item From the selected sub-population $j$, it samples a new parameter configuration from the multivariate Gaussian
    \[
        x \sim \mathcal{N}\bigl(m_j,\ \sigma_j^2 C_j\bigr).
    \]
\end{enumerate}

During a call to \texttt{update(params, score)}, the agent identifies which sub-population $j$ produced the parameters and updates its mean $m_j$, covariance matrix $C_j$, and global step-size $\sigma_j$ following the standard \cmaes update equations; see for example the original paper of Hansen and Ostermeier~\cite{hansen2001completely} for the detailed update rules. After each update, we apply a simple niching mechanism: the agent compares the means of all sub-populations; if two means become too close, the later sub-population is restarted.

The \texttt{time\_warp(steps)} function is identical to that of the Gaussian agent.

\section{Experiments}\label{sec:experiments}

We evaluate the different agents on two different generative systems. 

\vspace{1em}

\noindent \textbf{Superformula}
The first system is based on the \emph{superformula}, a generalization of the superellipse introduced by Johan Gielis in 2003~\cite{gielis2003generic}. This formula produces a wide range of organic shapes using only a small set of parameters. On top of the superformula, we also introduce additional noise and multiple rotated layers to create more diversity in the generated images. The parameter space has $7$ dimensions of type float and integer.

\vspace{1em}

\noindent \textbf{Suburbia}
The second system is Julien Gachadoat's generative art algorithm called \emph{Suburbia}. The writing of the algorithm started in 2012 with Processing and was regularly updated between 2018 and 2021 to create a large collection of plotted artworks that somehow has defined the style of Gachadoat. It was then adapted to \texttt{p5.js} in 2025 to be tested with \param. The algorithm relies on a grid of rectangles that are subdivived recursively, spawning polygons that are "decorated" with lines using various strategies of combinations and distributions prone to parametrization. The parameter space has $23$ dimensions of type float, integer and choice.

The qualitative experiment aims to visualize the trajectories followed by the different agents during their exploration. To this end, we run each agent for a fixed number of iterations and display representative drawings generated from each of the agent's internal modes at the end of the run. This provides a qualitative view of how individual modes correspond to distinct visual styles or motifs. We present here the result for the Open-ended agent; results for the other agents are very similar and are therefore omitted for brevity.

\begin{figure}[H]
        \centering
        \includegraphics[width=0.7\linewidth]{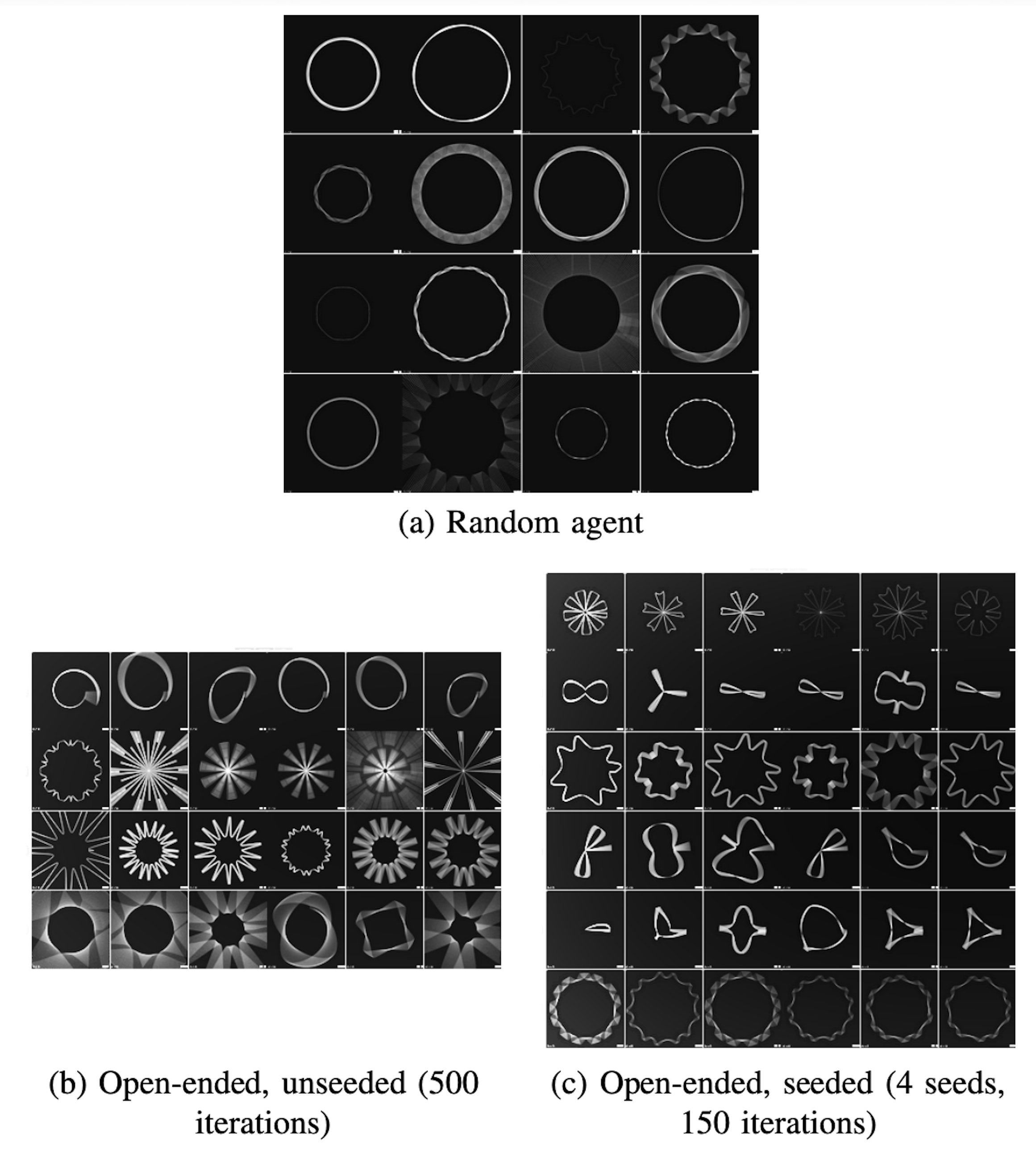}
    \caption{Comparison between several agents.}
    \label{fig:superformula_comparison}
\end{figure}

As a baseline, Figure~\ref{fig:superformula_comparison} (a) shows 16 images sampled by the random agent on the superformula system. Observe that the outputs are quite similar to one another.

Figure~\ref{fig:superformula_comparison} (b) and (c) illustrates the evolution of the Open-ended agent on the Superformula system, both without manual seeding (after 500 iterations, about 15 minutes of interaction) and with four manually chosen initial seeds (after 150 iterations). Each row corresponds to an internal mode of the agent and shows the six most recent outputs. The unseeded setting allows us to evaluate the agent's exploratory capacity on its own. Each row corresponds to an internal mode of the agent and shows its six most recent outputs. Without seeding, the agent must overcome an initial cold-start phase, whereas manual seeding leads to the rapid, almost immediate, emergence of visually interesting outputs. In both cases, feedback was provided using a minimal scheme: assigning the maximum score to selected images and leaving all others unrated.

\begin{figure*}[t]
    \centering
    \includegraphics[width=\textwidth]{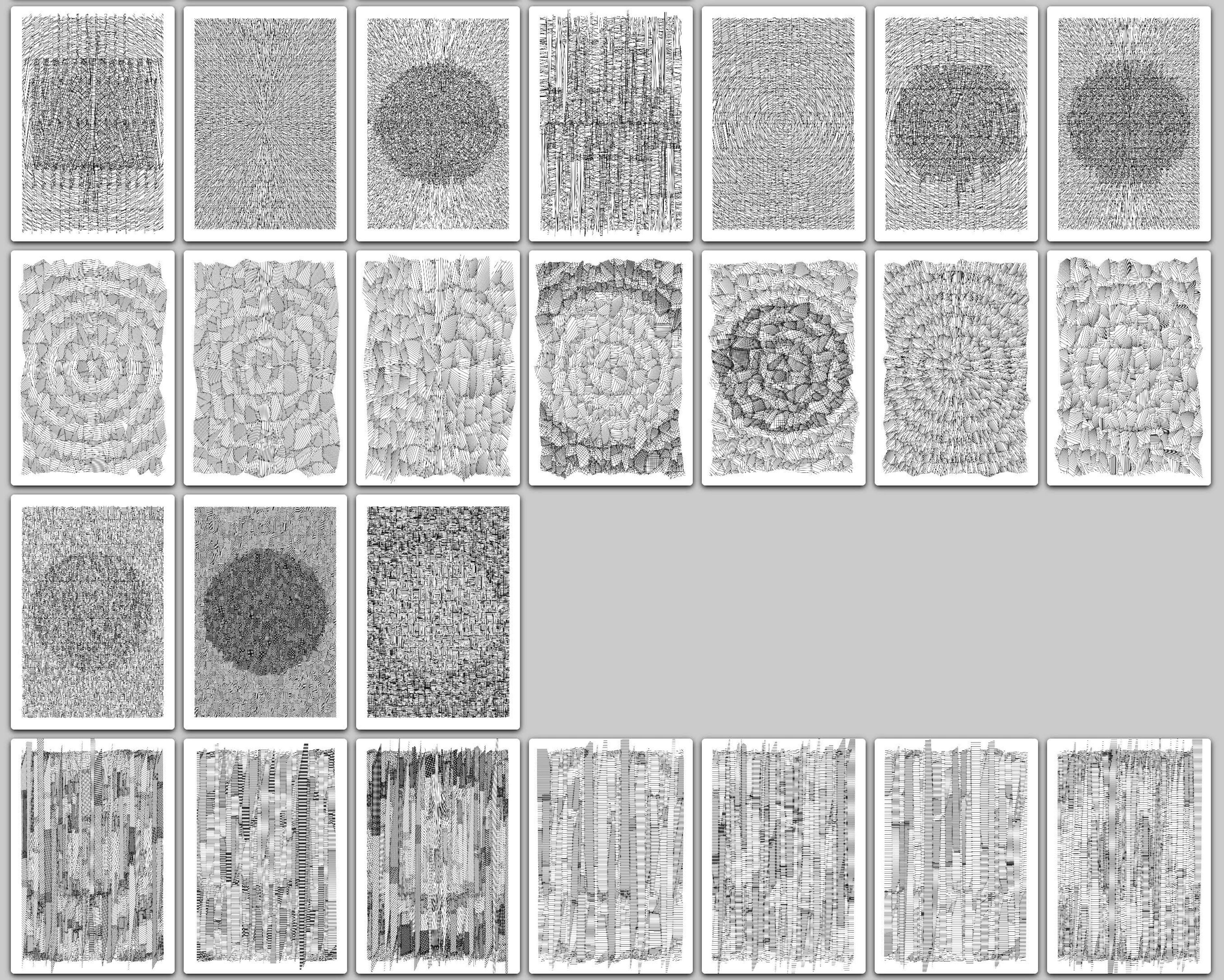}
    \caption{Evolution of the Open-ended agent on the Suburbia system after roughly 200 iterations. Each row corresponds to a different internal mode of the agent and shows the last images generated from that mode.}
    \label{fig:suburbia}
\end{figure*}

On the more complex Suburbia system, we observe similar trends. As an example, Figure~\ref{fig:suburbia} shows four of the fifteen different modes captured by the Open-ended agent after roughly 200 iterations.

\section{Future work}\label{sec:discussion}

A word on what comes next. We have privileged simplicity in the initial design of both the framework and the agents, and several directions for future work naturally follow from this first implementation.

A promising next step is to integrate more advanced exploration and niching strategies from multi-modal optimization. In particular hierchical methods, where search is organized across multiple levels of different granularities. Beyond their algorithmic value, such hierarchical structures are also meaningful from an artistic perspective: the hierarchy can be exposed to the user as a navigable diagram, allowing them to revisit, branch from, or refine any node in the creative process. This would turn the hierchical search tree itself into an interactive creative map.

Another direction is the integration of different forms of user feedback, including pairwise comparisons or qualitative annotations. Indeed, asking users to assign absolute scores to images can be cognitively demanding and inconsistent.

Finally, we believe it is important to release the framework to a broader audience and enable the artistic community to adopt, adapt, and improve the tool. To this end, we also aim to make the framework agnostic to backend techniques, so that users without technical expertise can employ it more effectively.

\newpage
\bibliographystyle{unsrt}
\bibliography{biblio}

@article{Molnar1975,
  title={Toward aesthetic guidelines for paintings with the aid of a computer},
  author={Molnar, Vera},
  journal={Leonardo},
  pages={185--189},
  year={1975},
  publisher={JSTOR}
}

@inproceedings{akiba2019optuna,
  title={Optuna: A next-generation hyperparameter optimization framework},
  author={Akiba, Takuya and Sano, Shotaro and Yanase, Toshihiko and Ohta, Takeru and Koyama, Masanori},
  booktitle={Proceedings of the 25th ACM SIGKDD international conference on knowledge discovery \& data mining},
  pages={2623--2631},
  year={2019}
}

@incollection{feurer2019hyperparameter,
  title={Hyperparameter optimization},
  author={Feurer, Matthias and Hutter, Frank},
  booktitle={Automated machine learning: Methods, systems, challenges},
  pages={3--33},
  year={2019},
  publisher={Springer International Publishing Cham}
}

@book{reas2007processing,
  title={Processing: a programming handbook for visual designers and artists},
  author={Reas, Casey and Fry, Ben},
  volume={6812},
  year={2007},
  publisher={Mit Press}
}

@inproceedings{processing2001,
  title={Processing: a learning environment for creating interactive web graphics},
  author={Reas, Casey and Fry, Benjamin},
  booktitle={ACM SIGGRAPH 2003 Web Graphics},
  pages={1--1},
  year={2003}
}

@book{mcCarthyp5js,
  title     = {Getting Started with p5.js},
  author    = {McCarthy, Lauren and Reas, Casey and Fry, Ben},
  publisher = {Make Community, LLC},
  year      = {2015}
}

@book{Migayrou2018CoderLeMonde,
  editor    = {Migayrou, Frédéric},
  title     = {Coder le monde: Mutations/Créations},
  publisher = {Éditions HYX},
  address   = {Orléans, France},
  year      = {2018},
  isbn      = {978-2-37382-011-9},
  note      = {Coédité avec les Éditions du Centre Pompidou},
}

@article{dreher2014history,
  title={History of Computer Art},
  author={Dreher, Thomas},
  year={2014},
  note={https://iasl.uni-muenchen.de/links/GCA-III.2e.html}
}

@book{maeda2001design,
  title={Design by numbers},
  author={Maeda, John},
  year={2001},
  publisher={MIT press}
}

@inproceedings{mccormack2022quality,
  title={Quality-diversity for aesthetic evolution},
  author={McCormack, Jon and Cruz Gambardella, Camilo},
  booktitle={International Conference on Computational Intelligence in Music, Sound, Art and Design (Part of EvoStar)},
  pages={369--384},
  year={2022},
  organization={Springer}
}

@article{ibarrola2022towards,
  title={Towards co-creative drawing based on contrastive language-image models},
  author={Ibarrola, Francisco and Bown, Oliver and Grace, Kazjon},
  journal={coordinates},
  volume={10},
  pages={2},
  year={2022}
}

@article{gielis2003generic,
  title={A generic geometric transformation that unifies a wide range of natural and abstract shapes},
  author={Gielis, Johan},
  journal={American journal of botany},
  volume={90},
  number={3},
  pages={333--338},
  year={2003},
  publisher={Wiley Online Library}
}

@article{hansen2001completely,
  title={Completely derandomized self-adaptation in evolution strategies},
  author={Hansen, Nikolaus and Ostermeier, Andreas},
  journal={Evolutionary computation},
  volume={9},
  number={2},
  pages={159--195},
  year={2001},
  publisher={MIT Press}
}

@article{li2016seeking  ,
  title={Seeking multiple solutions: An updated survey on niching methods and their applications},
  author={Li, Xiaodong and Epitropakis, Michael G and Deb, Kalyanmoy and Engelbrecht, Andries},
  journal={IEEE Transactions on Evolutionary Computation},
  volume={21},
  number={4},
  pages={518--538},
  year={2016},
  publisher={IEEE}
}

\end{document}